\documentclass[10pt,twocolumn,letterpaper]{article}

\usepackage{cvpr}              
\usepackage{xcolor}         
\usepackage{graphicx}
\usepackage{bbding}
\usepackage{amsmath}
\usepackage{multirow}
\usepackage{tabularx}
\definecolor{cvprblue}{rgb}{0.21,0.49,0.74}
\usepackage[pagebackref,breaklinks,colorlinks,allcolors=cvprblue]{hyperref}
\usepackage{lipsum}

\def\paperID{1910} 
\def\confName{CVPR}
\def\confYear{2025}

\title{Efficient Visual State Space Model
for Image Deblurring}

\author{%
    Lingshun Kong$^1$, Jiangxin Dong$^1$, Jinhui Tang$^1$, Ming-Hsuan Yang$^{2, 3}$, Jinshan Pan$^{1,\dagger}$\\
    $^1$Nanjing University of Science and Technology\\
    $^2${University of California, Merced} $^3$Google \\
}

\begin{document}
\maketitle

\newcommand\blfootnote[1]{%
\begingroup
\renewcommand\thefootnote{}\footnote{#1}%
\addtocounter{footnote}{-1}%
\endgroup
}
\def\thefootnote{$\dagger$}\footnotetext{Corresponding author}  

\begin{abstract}
Convolutional neural networks (CNNs) and Vision Transformers (ViTs) have achieved excellent performance in image restoration.
While ViTs generally outperform CNNs by effectively capturing long-range dependencies and input-specific characteristics, their computational complexity increases quadratically with image resolution. 
This limitation hampers their practical application in high-resolution image restoration.
In this paper, we propose a simple yet effective visual state space model (EVSSM) for image deblurring, leveraging the benefits of state space models (SSMs) for visual data. 
In contrast to existing methods that employ several fixed-direction scanning for feature extraction, which significantly increases the computational cost,
we develop an efficient visual scan block that applies various geometric transformations before each SSM-based module, capturing useful non-local information and maintaining high efficiency.
In addition, to more effectively capture and represent local information, we propose an efficient discriminative frequency domain-based feedforward network (EDFFN), which can effectively estimate useful frequency information for latent clear image restoration.
Extensive experimental results show that the proposed EVSSM performs favorably against state-of-the-art methods on benchmark datasets and real-world images. 
\end{abstract}

\section{Introduction}
Image deblurring aims to restore sharp images from blurry ones, which is drawing much attention due to the popularization of various cameras and handheld imaging devices.
This task is challenging, as only the blurred images are available while losing access to the blur and latent images.

Significant progress has been made due to the development of deep convolutional neural networks (CNNs)~\cite{SRN,SSN,MIMO,MPRNet,NAFNet}. 
However, as the main operation in CNNs, the convolution is spatially invariant and has limited receptive fields. This does not capture the spatially variant properties of the image contents and cannot explore non-local information that is beneficial for deblurring. 

In contrast to the convolution operation, the self-attention mechanism~\cite{Transformer} in Transformers can capture global information by computing correlations between each token and all other tokens, which can extract better features for image deblurring.
However, the self-attention mechanism (i.e., the scaled dot-product attention) entails quadratic space and time complexity regarding the number of tokens, which becomes unacceptable when handling high-resolution images.
Although the local window-based methods~\cite{Uformer,SwinIR}, transposed attention~\cite{Restormer}, and the frequency domain-based approximation~\cite{fftformer} have been developed to reduce the computational cost, these approaches sacrifice their abilities to model non-local information~\cite{Uformer,SwinIR} and spatial information~\cite{Restormer,fftformer}, which thus affects the quality of restored images. 
Therefore, it is of great need to develop an efficient approach that can explore non-local information for high-quality deblurring performance while not significantly increasing the computational cost. 
Recently, state space models (SSMs)~\cite{S4,LSSL} have demonstrated significant potential in modeling long-range dependencies for natural language processing (NLP) tasks with linear or near-linear computational complexity.
The improved SSM, specifically Mamba~\cite{mamba}, develops a selective scanning mechanism (S6) and can remember relevant information and ignore irrelevant content while achieving linear computational complexity. 
This motivates us to utilize SSMs to efficiently explore valuable non-local information for improved image deblurring.
However, since Mamba is designed to handle one-dimensional (1D) sequences, it requires first flattening of the image data to a 1D image sequence if Mamba is easily applied to visual tasks~\cite{ALGNet,CU-Mamba}.
This disrupts the spatial structure of the image, making it difficult to capture local information from various adjacent pixels. 
Several approaches adopt a multi-direction scan mechanism to utilize the state space model in visual applications~\cite{Vmamba,MambaIR,VmambaIR}.
However, the multi-direction scan mechanism significantly increases the computational cost.
In this paper, we propose an effective and efficient visual state space model for image deblurring.
We find that existing visual state space models mostly adopt several fixed-direction scanning for feature extraction, which may not explore non-local information adaptively and lead to higher computational costs.
We thus develop a simple yet effective scanning strategy that captures non-local spatial information while maintaining low computational costs.
Specifically, we only scan the input feature in one direction, but employ a simple geometric transformation before each scan, which effectively and adaptively explores useful information with a minimal increase in computational costs. 
%
In addition, we note that recent approaches based on the Mamba framework typically derive the parameters $ B $, $ C $, and $ \Delta $ using a purely linear transformation of the input features. 
This results in these parameters encoding identical spatial information, as they are all derived through linear transformations of the same input features. 
Consequently, this limits the model's ability to capture diverse spatial patterns, which is detrimental to the subsequent scanning process.
To overcome this, we introduce a 1D depth-wise convolution operation. This operation enables the model to capture both global patterns and local details, allowing the SSMs to focus on a broader range of spatial information and thereby enhancing their ability to capture fine spatial details.
Moreover, due to the presence of the geometric transformation we propose, one-dimensional convolutions can also aggregate information from the original 2D input.

Since SSMs primarily focus on global information, we propose an efficient discriminative frequency domain-based FFN (EDFFN) to enhance local details. 
The original DFFN \cite{fftformer} learns a quantization matrix in the frequency domain mid-way through the FFN to adaptively determine which frequency information should be preserved, but this results in a significant computational cost. 
Since the channel dimension of the features in the middle of the FFN is often several times that of the original input (e.g., the number of middle features of DFFN~\cite{fftformer} is three times that of the input), the computational cost becomes substantial when performing a fast Fourier transform on them.
Thus, we improve the original DFFN~\cite{fftformer} by learning the quantization matrix in the frequency domain at the end of the FFN, which greatly reduces the computational cost of the network.

The main contributions are summarized as follows.
\begin{itemize}
    \item We propose a simple yet effective visual state space model that efficiently restores high-quality images. Compared to existing methods based on SSM, our approach can efficiently restore blurred images with only a quarter of the computational cost.
    \item We propose an efficient discriminative frequency domain-based FFN that addresses the high computational cost issue of the previous frequency domain-based FFN, achieving a significant reduction in running time without compromising performance.
    \item We quantitatively and qualitatively evaluate the proposed method on benchmark datasets and real-world images and show that it performs effectively and efficiently against state-of-the-art methods.
\end{itemize}

\section{Related Work}

{\flushleft \textbf{Deep CNN-based image deblurring methods.}}
In recent years, significant progress has been made in image deblurring using deep CNN-based methods~\cite{GoPro,SRN,SSN,DMPHN,MPRNet,MIMO,NAFNet}. 
In~\cite{GoPro}, a deep CNN is proposed based on a multi-scale framework that directly estimates clear images from blurry ones. 
Tao et al.~\cite{SRN} introduce an effective scale recurrent network to enhance the utilization of information from each scale within the multi-scale framework. 
Zhang et al.~\cite{DMPHN} propose an effective network that adopts a multi-patch strategy for image deblurring. The deblurring process is executed step by step, enabling the network to refine the output progressively.
To further exploit the features extracted at different stages, Zamir et al.~\cite{MPRNet} introduce a cross-stage feature fusion technique, aiming to improve the overall performance of the deblurring method.
To mitigate the problem of high computational cost associated with multi-scale frameworks, Cho et al.~\cite{MIMO} present a multi-input and multi-output network architecture, reducing the computational burden while maintaining the deblurring performance.
Chen et al.~\cite{NAFNet} analyze the baseline modules and propose simplified versions to improve the efficiency of image restoration.
However, since the convolution operation is spatially invariant and spatially local, it cannot effectively model the global and spatially variable information, limiting its ability to achieve better image restoration.

{\flushleft \textbf{Transformer-based image deblurring methods.}}
Transformers can establish long-range dependencies and effectively model global information, 
thus researchers have extended its application to image super-resolution~\cite{SwinIR}, image deblurring~\cite{Restormer, fftformer}, and image denoising \cite{IPT, Uformer}.
The self-attention mechanism in Transformer requires quadratic computational complexity, which is unacceptable for image restoration tasks with high-resolution images.
To reduce the computational complexity of Transformers, Zamir et al. \cite{Restormer} propose an efficient Transformer model that computes scaled dot-product attention in the feature depth domain. 
Tsai et al. \cite{Stripformer} simplify the self-attention calculation by constructing intra- and inter-strip tokens to replace global attention.
Wang et al. \cite{Uformer} introduce a Transformer based on a UNet architecture that applies non-overlapping window-based self-attention for single image deblurring. 
Kong et al. ~\cite{fftformer} propose a frequency domain-based Transformer and achieve state-of-the-art results.
Although these methods employ various strategies to reduce computational complexity, they cannot characterize long-range dependencies and non-local information effectively.
In contrast, we develop an efficient visual state space model that explores useful non-local information with low computational costs.

{\flushleft \textbf{State space models.}}
State Space Models (SSMs) have been a time series analysis and modeling cornerstone for decades.
Recent methods~\cite{S4,LSSL} have adopted SSMs to capture long-range dependencies for sequence modeling.
SSM-based methods can be computed efficiently using recurrence or convolution, with linear or near-linear computational complexity.
Gu et al.~\cite{HiPPO} propose a framework (HiPPO) that offers the abstraction of the optimal function approximation in terms of time-varying measures.
The method~\cite{LSSL} develops a linear state-space layer to handle long-range dependency problems.
To address the issue of high computational and memory requirements induced by state representation, S4~\cite{S4} proposes to normalize parameters into a diagonal structure.
Furthermore, Mamba~\cite{mamba} introduces a selective scanning layer with dynamic weights, showcasing significant potential in natural language processing.
To apply SSMs for visual tasks, recent methods \cite{Vmamba,VmambaIR,MambaIR,wang2024lingen} adopt multi-direction scanning strategies, which will increase the computational cost.
In contrast, we propose an efficient visual scan block that employs a geometric transformation before each scan to achieve non-local information exploration with high efficiency.
\begin{figure*}[!t]
    \centering
 \includegraphics[width=0.98\textwidth]{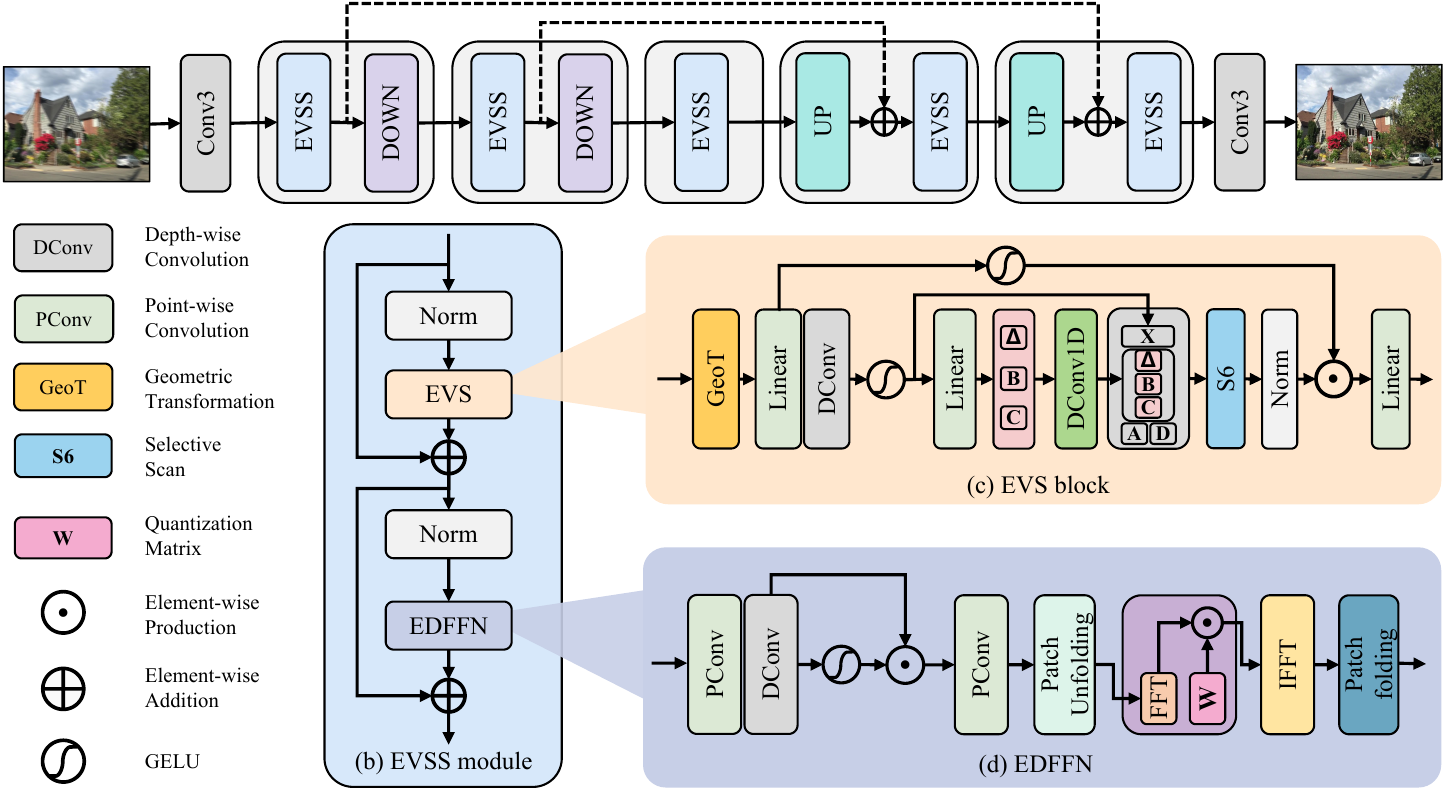}
 \vspace{-1mm}
 \caption{Efficient visual state space model. To efficiently restore high-quality images with SSMs, we propose an effective EVSS module that involves an efficient EVS block and an efficient EDFFN block. A geometric transformation is employed at the beginning of each EVS block to facilitate more useful information exploration in the following selective scan with a minimal increase in computational costs.}
 \vspace{-5mm}
 \label{fig: Network}
\end{figure*}

\section{Efficient Visual State Space Model}
\label{Methods}
Our goal is to present an effective and efficient method to explore the properties of state space models for high-quality image deblurring. 
To this end, we propose an efficient visual state space model (EVSSM) that can explore more non-local information for visual tasks with a minimal increase in computational costs.

\subsection{Overall architecture}
As shown in Figure~\ref{fig: Network}, the overall architecture of the proposed EVSSM is based on a hierarchical encoder-decoder framework~\cite{fftformer}.
Given a blurry image $I_{blur}\in\mathbb{R}^{H\times W\times 3}$, we first employ a  $3\times 3$ convolutional layer to obtain the shallow feature $F_s\in\mathbb{R}^{H\times W\times C}$, where $H\times W$ denotes
the spatial dimension and $C$ is the number of feature channels.
Then, the shallow feature $F_s$ is put into a 3-level symmetric encoder-decoder network.
The encoder/decoder at each level is composed of several efficient vision state space (EVSS) modules (see Section \ref{ssec:evss}).
For the encoder/decoder at level-$l$, the input feature is progressively processed by each EVSS module and the intermediate feature ${F^l_{enc}}$~/~${F^l_{dec}}$ $\in\mathbb{R}^{\frac{H}{2^{l-1}}\times \frac{W}{2^{l-1}}\times {2^{l-1}}C}$ ($l=1, 2, 3$ in this work) is generated.
We then employ bilinear interpolation and $1\times 1$ convolution to achieve upsampling and downsampling and add the skip connection between the encoder and decoder at each level.
Finally, a $3\times 3$ convolutional layer is applied to the feature ${F^3_{dec}}$ to generate the residual image $R\in\mathbb{R}^{H\times W\times 3}$.

The restored image $I_{deblur}$ is obtained by $I_{deblur} = R + I_{blur} = \mathcal{N}(I_{blur}) + I_{blur}$, and $\mathcal{N}$ denotes the proposed encoder-decoder network regularized by minimizing the following loss function:
\begin{equation}
\mathcal{L} = \lVert I_{deblur} - I_{gt} \rVert_1 + \lambda\lVert\mathcal{F}(I_{deblur}) - \mathcal{F}(I_{gt}) \rVert_1,\\
\label{eq: loss}
\end{equation}
where $\mathcal{F}$ denotes the discrete Fourier transform and the weight parameter $\lambda$ is empirically set as 0.1.

\subsection{Efficient vision state space module}
\label{ssec:evss}
{\flushleft \textbf{State space model.}}
A state space model is a mathematical framework commonly used in time series analysis and control systems. 
The state equation describes the evolution of an underlying system over time, representing the relationship between the hidden states of the system and their temporal dynamics.
The input signal $x(t)$ is assigned to the output response $y(t)$ through the hidden state $h(t)$.
It is typically modeled as a set of first-order difference or differential equations:
\begin{equation}
h^{'}(t) = Ah(t) + Bx(t), ~~~~ y(t) = Ch(t) + Dx(t),
\label{eq: ssm}
\end{equation}
where $A, B, C$, and $D$ are learnable weight matrices.
To this end, the state equation can be discretized using the zero-order hold (ZOH) technique:
\begin{equation}
\begin{split}
&  h_t = \Bar{A}h_{t-1} + \Bar{B}x_t,~~~~  y_t = Ch_t + Dx_t,\\
&  \Bar{A} = e^{\Delta A},~~~~ \Bar{B} = (\Delta A)^{-1}(e^{\Delta A}-I)\cdot \Delta B.
\end{split}
\label{eq: discrete SSM}
\end{equation}
Based on \eqref{eq: discrete SSM}, Mamba~\cite{mamba} proposes a selective scanning mechanism (S6) to achieve input-dependent weights and linear computational complexity simultaneously.
Using SSMs for NLP tasks poses no issue since natural language data is inherently a causal sequence. 
However, visual tasks present significant challenges as visual data is fundamentally non-sequential and contains spatial information such as local textures and global structures.
As S6 is a recursive process, when processing an input in the current timestep $t$, it can only utilize information from previous timesteps and not consider information from future timesteps.
{\flushleft \textbf{Efficient visual scan (EVS) block.}}
A straightforward approach to solving this problem is to scan the visual data in different directions (e.g., forward and backward).
However, this strategy increases the computational cost by a significant factor.
For example, the computational cost of the scanning in VMamba~\cite{Vmamba} is $4\times$ higher than that of Mamba~\cite{mamba} due to its strategy of performing bidirectional scanning in the longitudinal and transverse directions.
Then, a natural question is whether it can reasonably process visual data using a state space model with a minimal increase in computational costs.
The answer is YES. We develop an efficient vision state space model that explores more useful information with a minimal increase in computational costs.
The key is the proposed EVS block. Instead of scanning in multiple directions, we scan only in one direction and apply one geometric transformation (e.g., flip and transposition) to the input before each scan.
Due to the translation-invariant property of convolution, the geometric transformation does not affect the convolution itself, but only influences the process of selective scanning.
Specifically, for each EVS block, assuming that it is in the $i$-th EVSS module of the whole network, we first transpose or flip the input feature $F_{in}$ as:
\begin{equation}
\begin{split}
&  G = 
\begin{cases}
  Transpose(F_{in}) & \text{if } i ~\% ~2 = 0, \\
  Flip(F_{in}) & \text{if } i ~\% ~2 = 1. \\
\end{cases}
\label{eq: EVSS}
\end{split}
\end{equation}
Here, we flip along both the horizontal and vertical axes of the feature in $Flip(\cdot)$.
According to \eqref{eq: EVSS}, the image feature will be automatically restored to the original spatial structure after every 4 EVSS modules.
In particular, if the total number of EVSS modules in our network is not divisible by 4, we can restore the original spatial structure by applying the corresponding inverse transformations, as both flip and transposition are reversible.
Recent approaches based on the Mamba framework typically derive the parameters $ B $, $ C $, and $ \Delta $ in a purely linear fashion from the input features. 
These parameters share identical spatial information, differing only along the channel dimension. 
Inspired by previous works on transformers, which effectively capture spatial dependencies by characterizing local information before computing self-attention, we propose a simple yet effective improvement.
Specifically, after the linear projection layer, we apply a 1D depth-wise convolution with a kernel size of 7 to each of $ B $, $ C $, and $ \Delta $. 
This allows our model to incorporate spatial context when deriving these parameters, thereby enriching their representation and enabling them to better capture local and global dependencies within the input data.
Specifically, due to the incorporation of geometric transformations, although we apply 1D convolutions on one-dimensional sequences, the model is capable of capturing not only unidirectional information but also multidirectional information from the 2D input. 
This enables the capture of richer contextual features across various orientations within the 1D space.

In this way, our EVS module effectively solves the above question, avoiding any additional computational burden other than efficient geometric transformations.
Then, the selective scan can be formulated as follows:
\begin{equation}
\begin{split}
&X_1,X_2 = \mathrm{split}(\mathrm{Linear}(G))\\
&X_1 = \sigma(\mathrm{Dconv_{3\times 3}}(X_1))\\
&[\Delta, B, C] = \mathrm{DConv1D}(\mathrm{Linear}(X_1))\\
&\hat{X_1} = \mathrm{S6}(X_1,A,B,C,D,\Delta)\\
&F_{out} = \mathrm{Linear}((\mathcal{L}(\hat{X_1})) \cdot \sigma(X_2)),
\label{eq: scan}
\end{split}
\end{equation}
where $\mathrm{Linear}(\cdot)$ is a fully connected layer, 
$\mathrm{DConv1D}(\cdot)$ and $\mathrm{DConv}_{3\times 3}(\cdot)$ indicate depth-wise convolutional layers with the filter size of $1\times1$ and $3\times 3$ pixels, respectively,
$\mathcal{L}(\cdot)$ denotes a normalization layer, $\mathrm{split}(\cdot)$ splits the image features in the channel dimension, $\sigma$ denotes the GeLU activation, S6 denotes the selective scanning mechanism proposed by~\cite{mamba},
$\cdot$ is the element-wise multiplication.
Figure~\ref{fig: Network}(c) shows the detailed network architecture.
{\flushleft \textbf{Efficient discriminative frequency domain-based FFN (EDFFN).}}
SSM primarily focuses on global information, but for image restoration, local information is equally important. In order to better characterize local information to assist in image restoration, we develop an efficient discriminative frequency domain-based FFN.
The FFN part is typically the core component of deep learning models, which can help the latent clear image reconstruction~\cite{Restormer,fftformer}.
FFTformer~\cite{fftformer} develops a discriminative frequency domain-based FFN (DFFN) that adaptively determines which frequency information should be preserved.
%
%
However, it is time-consuming to apply the Fast Fourier Transform (FFT) to the features, especially when the number of channels of the feature is large.
In contrast to DFFN that applies FFT in the middle of the FFN network, our approach performs frequency-domain screening on the features at the final stage of the FFN network using a smaller learnable quantization matrix W (as shown in Figure~\ref{fig: Network}(d)), thereby significantly reducing computational costs.
%
%

\section{ Experimental Results}
\label{experiments}

\subsection{ Datasets and implementation}
{\flushleft \textbf{Datasets.}}
Following existing state-of-the-art methods, we evaluate our approach on the commonly used GoPro dataset by Nah et al.~\cite{GoPro}, HIDE dataset by Shen et al.~\cite{HIDE}, and RealBlur dataset by Rim et al.~\cite{Realblur}.
The GoPro dataset contains $2103$ images for training and $1111$ images for testing.
The HIDE dataset includes $2025$ images mainly about humans for testing. 
The RealBlur dataset~\cite{Realblur} contains RealBlur-J and RealBlur-R subsets generated by different post-processing strategies. It has $3758$ images for training and $980$ images for testing.
For fair comparisons, we follow the protocols of these datasets to evaluate our method.
{\flushleft \textbf{Implementation details.}}
The shallow feature $F_s$ has a channel number of 48 and the numbers of EVSS modules in the encoder/decoder from level-$1$ to level-$3$ are [6, 6, 12].
We use the AdamW optimizer with default parameters in the training process.
We use the data augmentation method with the flipping and rotation operations to generate training data.
We apply progressive training similar to but simpler than \cite{Restormer}: the training starts with the patch size of $128 \times 128$ pixels and the batch size of $64$ for $300, 000$ iterations, where the learning rate gradually reduces from $1\times 10^{-3}$ to $1\times 10^{-7}$. 
Then the patch size is enlarged to $256\times256$ pixels with $16$ batches for $300, 000$ iterations where the learning rate is initialized as $5\times 10^{-4}$ and decreases to $1\times 10^{-7}$.
The learning rate is updated based on the Cosine Annealing scheme.
Unless otherwise specified, all experiments are conducted with the PyTorch framework on NVIDIA RTX 4090 GPUs.
The training code and models are available at \href{https://github.com/kkkls/EVSSM}{EVSSM}.
\begin{table}[!t]\footnotesize
\centering
    \caption{Quantitative evaluations on the GoPro
    dataset~\cite{GoPro} in terms of PSNR and SSIM.}
    \label{tab:gopro}
    \vspace{-2mm}
    \begin{tabular}{l@{}c@{\hspace{5mm}}c@{\hspace{5mm}}c@{\hspace{5mm}}c@{\hspace{3mm}}c@{}}
        \toprule
                & \multicolumn{2}{c}{GoPro} &   \multicolumn{2}{c}{HIDE}                    \\
        Methods                         & PSNR (dB)  &  SSIM & PSNR (dB) & SSIM \\
        \midrule
        SRN~\cite{SRN}                       & 30.26 & 0.9342 & 28.36 & 0.9040  \\
        DMPHN~\cite{DMPHN}                   & 31.20 & 0.9453 & 29.09 & 0.9240\\
        SAPHN~\cite{SAPHN}                   & 31.85 & 0.9480 & 29.98 & 0.9300 \\
        MIMO-Unet+~\cite{MIMO}               & 32.45 & 0.9567 & 29.99 & 0.9304 \\
        MPRNet~\cite{MPRNet}                 & 32.66 & 0.9589 & 30.96 & 0.9397 \\
        Restormer~\cite{Restormer}           & 32.92 & 0.9611 & 31.19 & 0.9418 \\
        Uformer-B~\cite{Uformer}             & 33.06 & 0.9670 & 30.90 & 0.9530 \\
        Stripformer~\cite{Stripformer}       & 33.08 & 0.9624 & 31.03 & 0.9395 \\
        DeepRFT+~\cite{Deeprft}              & 33.23 & 0.9632 & 31.42 & 0.9440 \\
        MPRNet-local~\cite{TLC}              & 33.31 & 0.9637 & 31.19 & 0.9418 \\
        Restormer-local~\cite{TLC}           & 33.57 & 0.9656 & 31.49 & 0.9447 \\
        NAFNet~\cite{NAFNet}                 & 33.71 & 0.9668 & 31.31 & 0.9427 \\
        GRL~\cite{GRL}                       & 33.93 & 0.9680 & 31.65 & 0.9470 \\
        FFTformer~\cite{fftformer}           & 34.21 & 0.9692 & 31.62 & 0.9455 \\
        CU-mamba~\cite{CU-Mamba}             & 33.53 & 0.9650 & 31.47 & 0.9440 \\
        MLWNet-B~\cite{MLWNet}               & 33.83 & 0.9676 & 31.06 & 0.9320      \\
        EVSSM                        & \textbf{34.51} & \textbf{0.9713} & \textbf{31.99} & \textbf{0.9503} \\
        \bottomrule
    \end{tabular}
   \vspace{-2mm}
\end{table}

\begin{table}[!t]\footnotesize
\centering
    \caption{Quantitative evaluations on the RealBlur
    dataset~\cite{Realblur} in terms of PSNR and SSIM.}
    \label{tab:realblur}
    \vspace{-2mm}
    \begin{tabular}{l@{}c@{\hspace{5mm}}c@{\hspace{5mm}}c@{\hspace{5mm}}c@{\hspace{2mm}}c@{}}
        \toprule
                & \multicolumn{2}{c}{Realblur-R} &   \multicolumn{2}{c}{Realblur-J}                    \\
        Methods                         & PSNR (dB)  &  SSIM & PSNR (dB) & SSIM \\
        \midrule
        DeblurGAN-v2~\cite{DeblurGANv2} & 36.44 & 0.9347 & 29.69 & 0.8703 \\
        SRN~\cite{SRN}                  & 38.65 & 0.9652 & 31.38 & 0.9091 \\
        MIMO-Unet+~\cite{MIMO}          &   -   &   -    & 31.92 & 0.9190 \\
        BANet~\cite{BANet}              & 39.55 & 0.9710 & 32.00 & 0.9230 \\
        DeepRFT+~\cite{Deeprft}         & 39.84 & 0.9721 & 32.19 & 0.9305 \\
        Stripformer~\cite{Stripformer}  & 39.84 & 0.9737 & 32.48 & 0.9290 \\
        FFTformer~\cite{fftformer}      & 40.11 & 0.9753 & 32.62 & 0.9326 \\
        MLWNet~\cite{MLWNet}            & 40.69 & 0.9755  & 33.84 & 0.9405   \\
        
        EVSSM                           & \textbf{41.27} & \textbf{0.9776} & \textbf{34.34} & \textbf{0.9456} \\
        \bottomrule
    \end{tabular}
   \vspace{-4mm}
\end{table}

\begin{figure*}[!t]
\footnotesize
\centering
\begin{tabular}{@{}c@{\hspace{0.5mm}}c@{\hspace{0.5mm}}c@{\hspace{0.5mm}}c@{}}
\includegraphics[width = 0.245\linewidth]{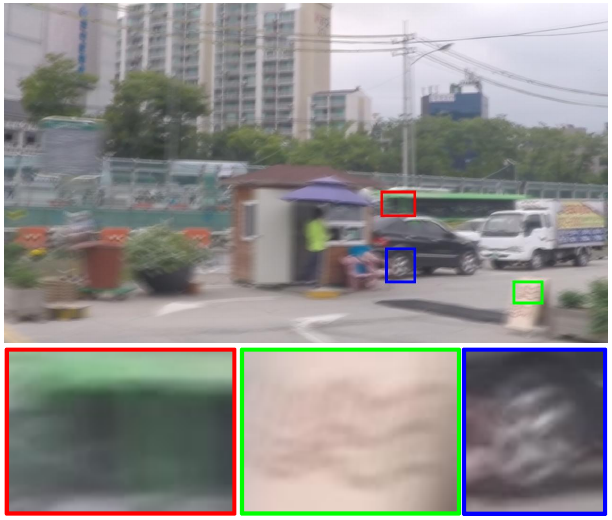}& 
\includegraphics[width = 0.245\linewidth]{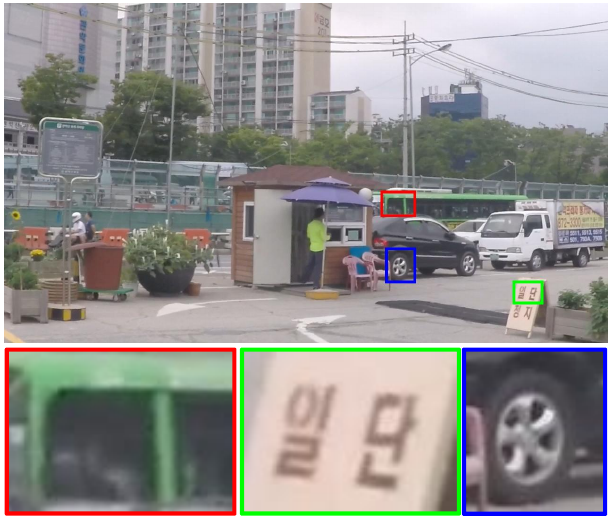}& 
\includegraphics[width = 0.245\linewidth]{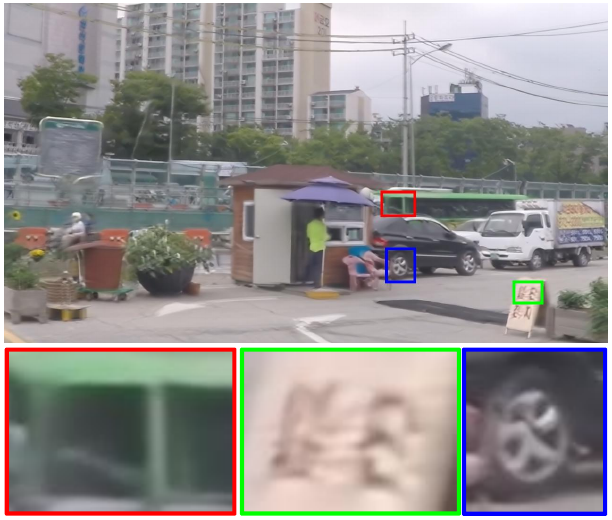}& 
\includegraphics[width = 0.245\linewidth]{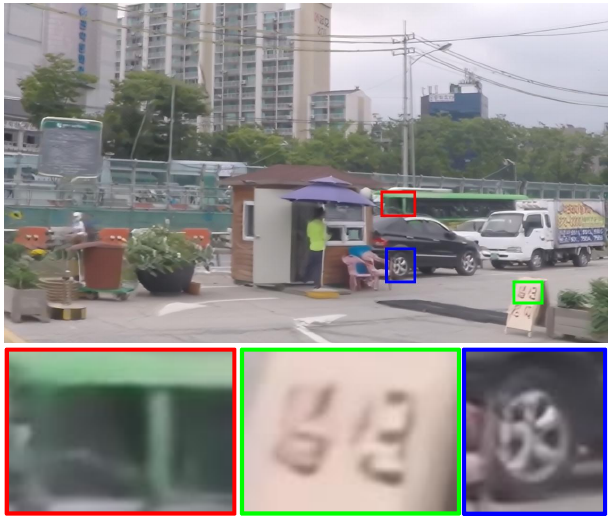}\\
(a) Blurred input &  (b) GT &  (c) MIMO-UNet+~\cite{MIMO}  &  (d) Stripformer~\cite{Stripformer}\\
\includegraphics[width=0.245\linewidth]{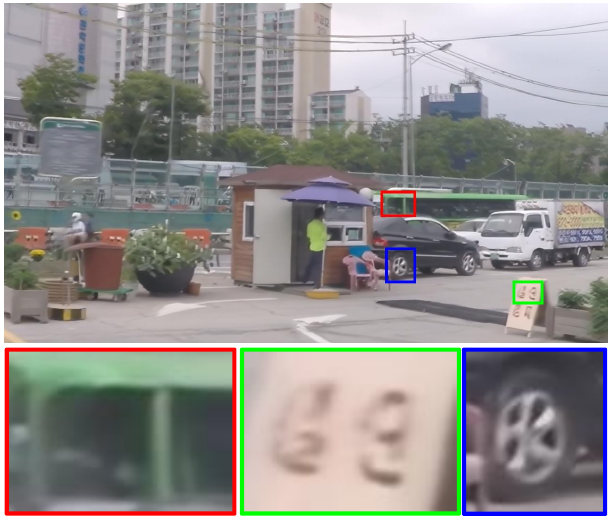}&
\includegraphics[width=0.245\linewidth]{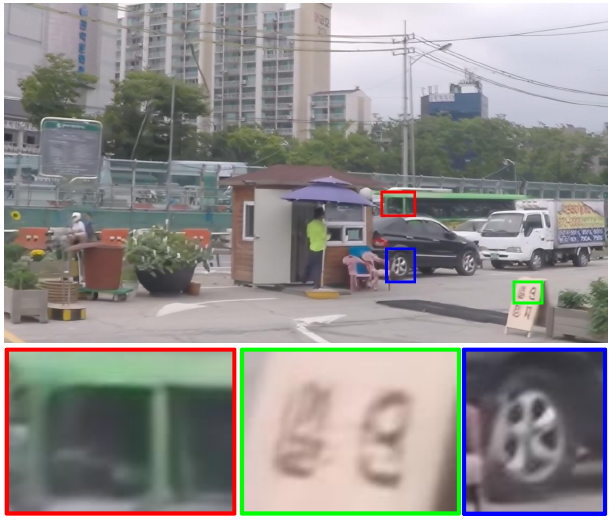}&
\includegraphics[width=0.245\linewidth]{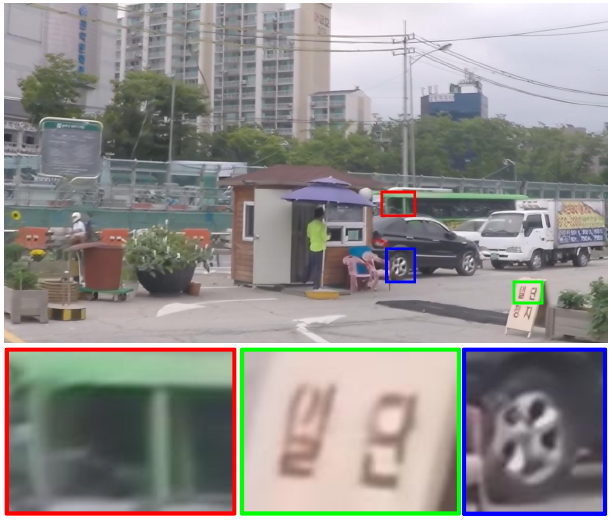}&
\includegraphics[width=0.245\linewidth]{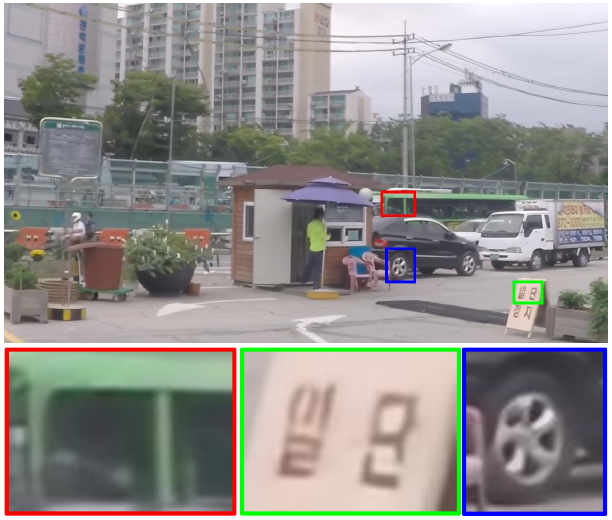}\\
(e) Restormer-local~\cite{TLC}& (f) NAFNet~\cite{NAFNet}&  (g) MLWNet~\cite{MLWNet}&  (h) EVSSM (ours) \\
\end{tabular}
\vspace{-3mm}
\caption{Deblurred results on the GoPro dataset~\cite{GoPro}. Compared to the results in (c)-(g), the proposed method generates a better deburred image with clearer structures and detail in (h).
}
\label{fig: result-gopro}
\vspace{-7mm}
\end{figure*}

\subsection{Comparisons with the state of the art}
{\flushleft \textbf{Evaluations on the GoPro dataset.}}
We first evaluate the performance of the proposed approach on the GoPro dataset~\cite{GoPro}.
We compare our method against state-of-the-art approaches, including CNN-based ones (SRN~\cite{SRN}, DMPHN~\cite{DMPHN}, MIMO-UNet+~\cite{MIMO}, MPRNet~\cite{MPRNet}, NAFNet~\cite{NAFNet}), Transformer-based ones (Uformer~\cite{Uformer}, Restormer~\cite{Restormer}, Stripformer~\cite{Stripformer}, Restormer-local~\cite{TLC}, GRL~\cite{GRL}, and FFTformer~\cite{fftformer}), MLP-based ones (MAXIM~\cite{maxim}), and SSM-based ones (CU-mamba~\cite{CU-Mamba}).
We retrain or fine-tune the deep learning-based methods for fair comparisons if they are not trained on the benchmarks.
As CU-mamba~\cite{CU-Mamba} does not provide training and test codes, for fair comparison, we compare with their reported results~\cite{CU-Mamba}.
We use the PSNR and SSIM as the evaluation metrics to measure the quality of each restored image.
Table~\ref{tab:gopro} shows the quantitative evaluation results, where our approach outperforms other methods with the highest PSNR and SSIM values.
%

\begin{figure*}[!t]
\footnotesize
\centering
\begin{tabular}{@{}c@{\hspace{0.5mm}}c@{\hspace{0.5mm}}c@{\hspace{0.5mm}}c@{}}
\includegraphics[width=0.245\linewidth]{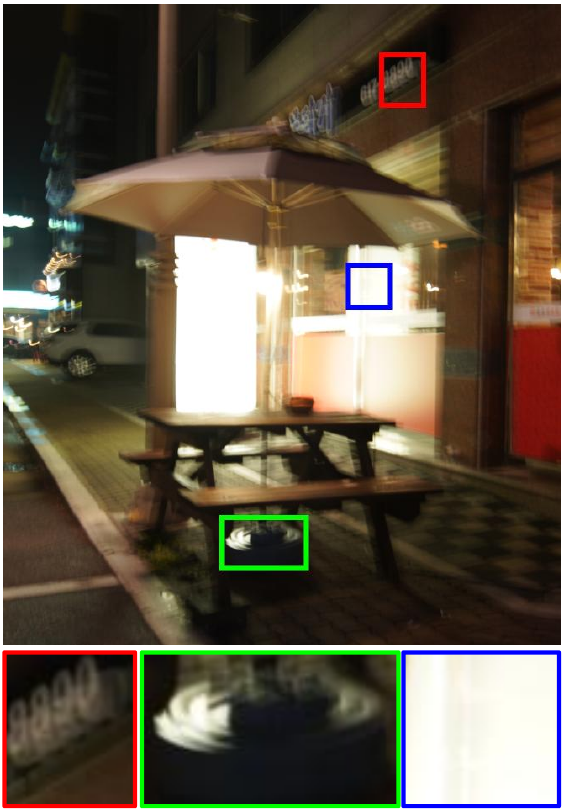} &
\includegraphics[width=0.245\linewidth]{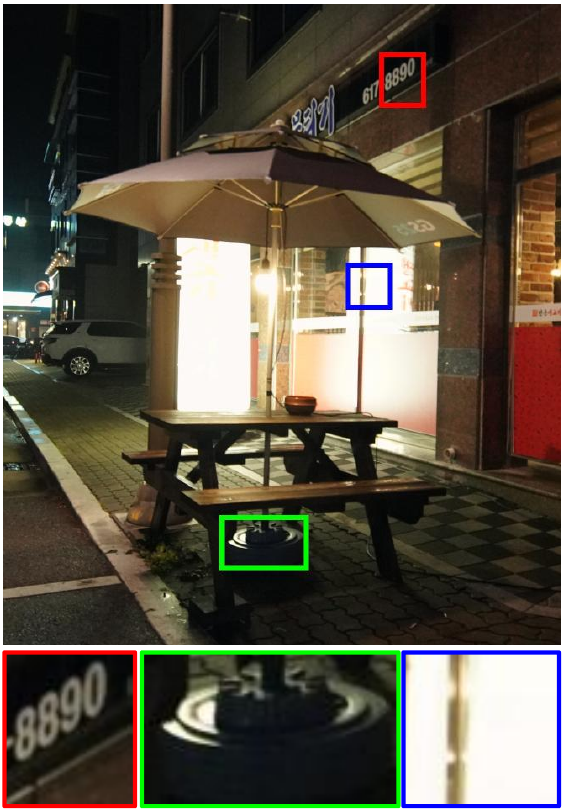} &
\includegraphics[width=0.245\linewidth]{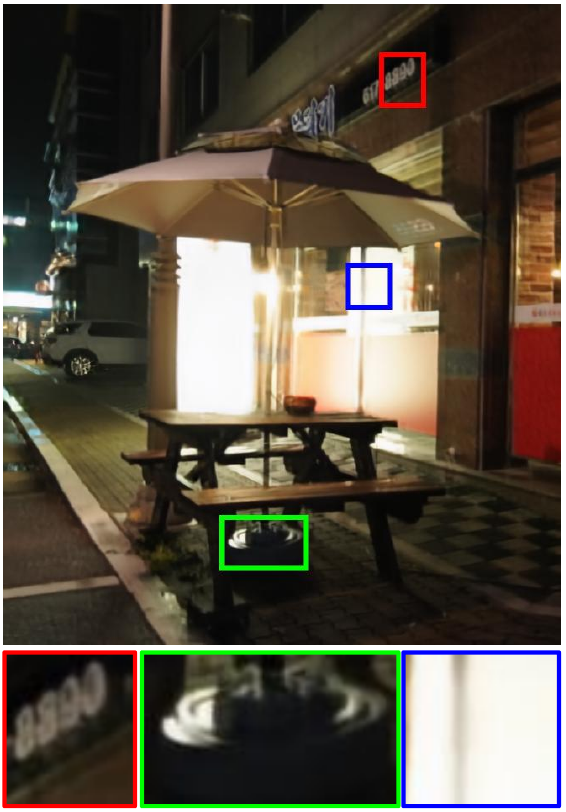} &
\includegraphics[width=0.245\linewidth]{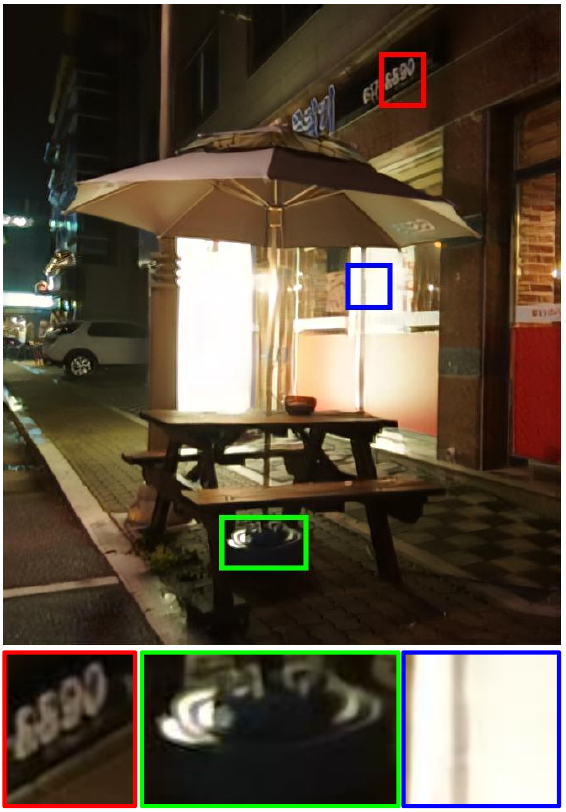} \\
(a) Blurred input &  (b) GT &  (c) SRN~\cite{SRN}  &  (d) MIMO-UNet+~\cite{MIMO}\\
\includegraphics[width=0.245\linewidth]{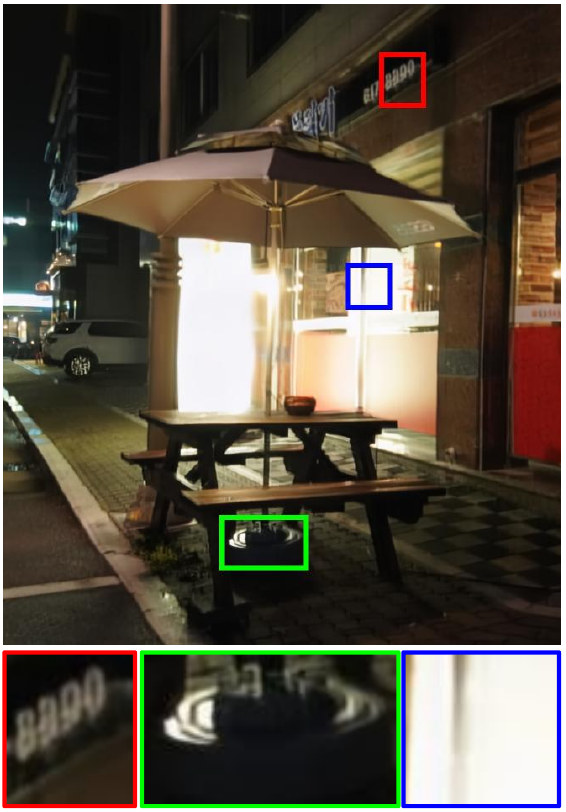}  &
\includegraphics[width=0.245\linewidth]{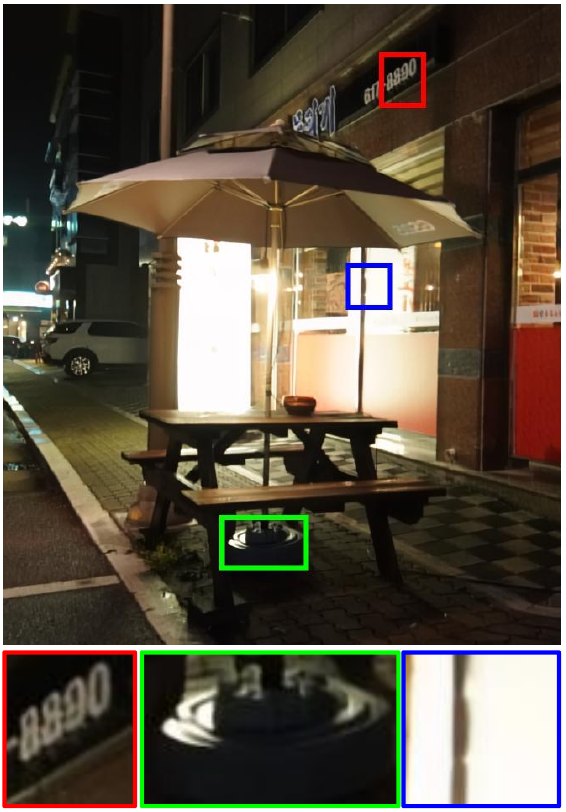} &
\includegraphics[width=0.245\linewidth]{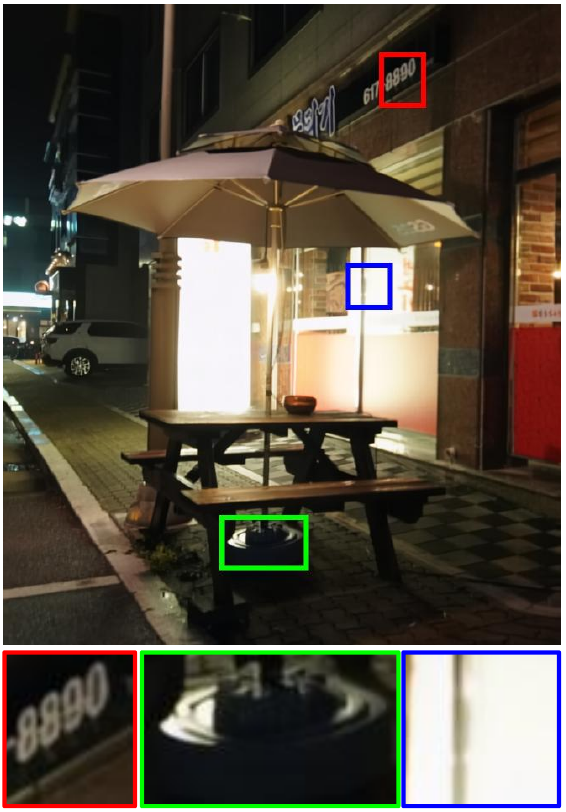} &
\includegraphics[width=0.245\linewidth]{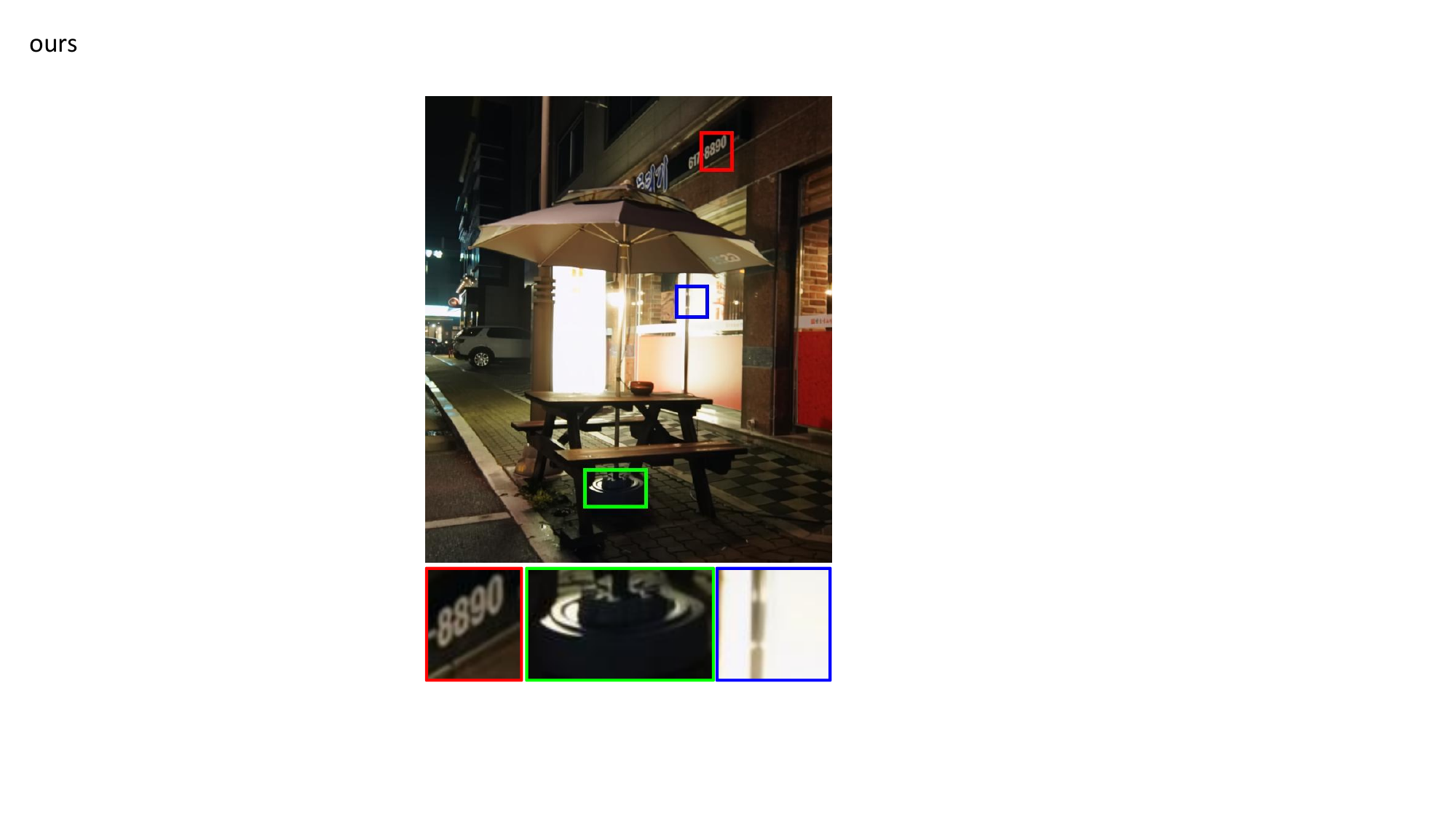} \\
(e) Stripformer~\cite{Stripformer}&  (f) FFTformer~\cite{fftformer}& (g) MLWNet~\cite{MLWNet}&   (h) EVSSM (ours) \\
\end{tabular}
\vspace{-3mm}

\caption{Deblurred results on the RealBlur dataset~\cite{Realblur}. The results obtained by~\cite{SRN,DeblurGANv2,MIMO} in (c)-(e) are still blurry. For other results by~\cite{Stripformer,fftformer} in (f)-(g), fine-scale structures are not effectively recovered. In contrast, our method restores a better-deblurred image with clearer characters.}%
\label{fig: result-realblur}
\vspace{-5mm}
\end{figure*}

Figure~\ref{fig: result-gopro} shows the evaluation results on the GoPro dataset~\cite{GoPro}.
Since CNN-based methods cannot effectively utilize global information, the images restored by~\cite{MIMO,NAFNet} still contain severe blur residuals as shown in Figure~\ref{fig: result-gopro}(c) and (f).
Although Transformer-based methods can model the global context, the methods~\cite{Stripformer,TLC,fftformer} have adopted various approximations to reduce the computational cost.
This affects their ability to model the long-range
information, and as a result, some main structures, {e.g.}, tire, and wheel, are not restored well (see Figure~\ref{fig: result-gopro}(d) and (e)).
In contrast to existing Transformer-based methods, we propose a simple yet effective visual state space model for image deblurring, which is effective at exploring non-local information with low computational costs. 
As shown in Figure~\ref{fig: result-gopro}(h), our method produces clearer results, especially for text and tire.
{\flushleft \textbf{Evaluations on the RealBlur dataset.}}
Using the same protocols, we evaluate our approach on the real-world blurry dataset by Rim et al.~\cite{Realblur}.
Table~\ref{tab:realblur} shows that the proposed method outperforms previous work significantly, improving the PSNR by at least $0.58$dB and $0.5$dB on the datasets of RealBlur-J and RealBlur-R, respectively.
Figure~\ref{fig: result-realblur} shows the visual comparisons, where our method generates the result with clearer characters and finer structural details.
%
\begin{table}[!t]\footnotesize
\centering
    \caption{Quantitative evaluations of the proposed method against state-of-the-art ones on the real-world dataset~\cite{SPA} for image deraining in terms of PSNR and SSIM.}
    \vspace{-3mm}

    \label{tab:SPA}
    \setlength{\tabcolsep}{2pt}

    \begin{tabular}{l@{}c@{\hspace{1mm}}c@{\hspace{1mm}}c@{\hspace{1mm}}c@{\hspace{-0mm}}c@{}}
        \toprule
       Method  & MSPFN~\cite{MSPFN} & RCDNet~\cite{RCDNet} & DualGCN~\cite{DualGCN}   &SPDNet~\cite{SPDNet}                \\
        \midrule 
        PSNR (dB)          & 43.43             & 43.36             & 44.18            & 43.20               
                       \\
        SSIM       & 0.9843            & 0.9831            & 0.9902                    & 0.9871    
                        \\
       \toprule
       Method          &Restormer~\cite{Restormer} &IDT~\cite{IDT} &DRSformer~\cite{DRSformer}  &EVSSM (ours)\\
        \midrule 
        PSNR (dB)           & 47.98               &  47.35        & 48.54        &\bf{49.00}     \\
        SSIM                & 0.9921             & 0.9930         &  0.9924      &\bf{0.9954}   \\
        \bottomrule
    \end{tabular}
    \vspace{-1mm}
\end{table}

\begin{table}[!t]\footnotesize
\centering
    \caption{Quantitative evaluations of the proposed method against state-of-the-art ones on the RESIDE-6K dataset~\cite{RESIDE} for image dehazing in terms of PSNR and SSIM.}
    \vspace{-3mm}

    \label{tab:RESIDE-6K}
    \setlength{\tabcolsep}{2pt} 

    \begin{tabular}{l@{}c@{\hspace{-0mm}}c@{\hspace{-0mm}}c@{\hspace{-0mm}}c@{\hspace{-0mm}}c@{}}
        \toprule
       Method  & MSCNN~\cite{MSCNN} &GFN~\cite{GFN}&MSBDN~\cite{MSBDN} & PFDN~\cite{PFDN} \\
        \midrule 
        PSNR (dB)   & 20.31            & 23.52         & 28.56            & 28.15           \\
        SSIM   & 0.863         & 0.905    & 0.966        & 0.962       \\     
               \toprule
       Method      &FFA-Net~\cite{FFA-Net}&AECRNet~\cite{AECRNet} &DehazeFormer~\cite{DehazeFormer}  &EVSSM (ours)\\
        \midrule 
        PSNR (dB)           & 29.96             & 28.52                  & 31.45            &  \bf{32.05}  \\
        SSIM                & 0.973  & 0.964                & \textbf{0.980}             & \textbf{0.980}\\
        \bottomrule
    \end{tabular}
    \vspace{-5mm}
\end{table}



%
{\flushleft \textbf{Evaluations on the HIDE dataset.}}
We examine our method on the HIDE dataset~\cite{HIDE} in Table~\ref{tab:gopro}. 
Similar to state-of-the-art methods~\cite{fftformer,TLC}, we directly use the models trained on the GoPro dataset for testing. 
The PSNR value of our approach is $0.35$dB higher than those from competing methods, showing that our method generalizes well, as the models are not trained on this dataset.
%

%
{\flushleft \textbf{Evaluations on datasets for other image restoration tasks.}}
We also evaluate our method on the real-world dataset~\cite{SPA} for image deraining and RESIDE-6K dataset~\cite{RESIDE} for image dehazing. 
Tables~\ref{tab:SPA}-\ref{tab:RESIDE-6K} show that the proposed approach performs favorably against state-of-the-art methods on image deraining and dehazing datasets, which demonstrates the generalization ability of our method on other image restoration tasks.

\section{Analysis and Discussion}
We have shown that the proposed efficient visual state space model generates favorable results compared to state-of-the-art methods. 
In this section, we provide more analysis of the proposed method and discuss the
effect of the main components. 
For the ablation studies in this section, we train our approach and all the baselines on the GoPro dataset using a batch size of $64$ and a patch size of $128 \times 128$ pixels for 300K iterations with the initial learning rate $1\times 10^{-3}$ gradually reduced to $1\times 10^{-6}$ with the cosine annealing schedule.
{\flushleft \textbf{Effectiveness of the EVS block.}}
The core of our proposed EVS block is the geometric transformation \eqref{eq: EVSS} applied at the beginning of this block.
To demonstrate the effect of the EVS block, we first remove the geometric transformation (\emph{one-direction} for short, which performs scanning in one direction as in~\cite{mamba}) and train this baseline model using the same settings as ours.
Table~\ref{tab: effect-of-EVSS} shows that the PSNR of our approach is $0.26$dB higher than this baseline method. 
The state space model requires flattening the image feature into a one-dimensional sequence, which compromises the spatial structural information of the visual data.
Compared to the baseline, our method with the geometric transformation can better explore non-local information.
Meanwhile, the number of parameters and the FLOPs of our method are the same as those of the baseline, and their runtimes are also almost identical, which demonstrates the effectiveness of the proposed EVS block in improving the ability of the state space model for handling visual data with a minimal increase in computational costs.
%

\begin{table}[!t]\footnotesize
  \caption{Effectiveness of the proposed EVS block, evaluated on the GoPro dataset~\cite{GoPro}.
  }
    \vspace{-3mm}
   \label{tab: effect-of-EVSS}
\footnotesize
 \centering
 \footnotesize
 \begin{tabular}{l|cc|r|c} 
    \toprule

   Scanning mode&Params & FLOPs &Runtime   & {PSNR (dB)/SSIM}\\
\hline
 one-direction   & 17.1M & 126G & 87.9ms   &  33.89/0.9671         \\
 two-direction   & 18.3M & 135G & 122.7ms  &  33.08/ 0.9687        \\
 four-direction  & 20.1M & 148G & 182.6ms  &  33.05/ 0.9686         \\
 EVS (ours)      & 17.1M & 126G & 88.8ms   &  \bf{34.15/0.9690}           \\
 \bottomrule
  \end{tabular}
  \vspace{-3mm}

\end{table}


In addition, we compare with two baselines that respectively perform scanning in two directions (\emph{two-direction} for short) and four directions~\cite{Vmamba} (\emph{four-direction} for short).
As Table~\ref{tab: effect-of-EVSS} shows, although scanning in multi-directions can alleviate the limitation of the state space model on handling visual data, it leads to an increase in the number of network parameters and the computational complexity, resulting in significantly longer runtimes, cf. $122.7$ms for one-direction \emph{vs.} $182.6$ms for four-direction \emph{vs.} $88.7$ms for our approach.
Note that the results of scanning in four directions are slightly lower than those of scanning in two directions. 
This is because the methods that scan in multiple directions simply employ the summation followed by normalization to fuse features extracted from different scanning directions.
Thus, the multi-directional information is not effectively and fully utilized.
In contrast to simultaneously scanning in multiple directions, our approach applies geometric transformations to the input feature at the beginning of each EVS block. 
This allows each scan to capture contextual information from different directions and mitigates the increase in computational complexity and runtime.
%


\begin{table}[!t]
  \caption{Effectiveness of the geometric transformation, evaluated on the GoPro dataset~\cite{GoPro}.
  }
  \vspace{-3mm}

   \label{tab: effect-of-scan}
\footnotesize
 \centering
 
 \begin{tabular}{lcc|c|cc}
    \toprule
   Method   &  Params  & FLOPs   & Runtime    & PSNR (dB)/SSIM\\
    \hline
 w/o F\&T       &17.1M &126G   &  87.9ms  &  33.89/0.9671 \\
  w/o F         &17.1M &126G   &  88.4ms  &  34.05/0.9685 \\
  w/o T         &17.1M &126G   &  88.4ms  &  34.03/0.9684  \\
  EVSSM (ours)  &17.1M &126G   &  88.8ms &\bf{34.15/0.9690} \\
 \bottomrule
  \end{tabular}

\vspace{-5mm}
\end{table}
%

%
%
\vspace{-2mm}
{\flushleft \textbf{Effectiveness of the geometric transformation.}} In the EVS block, we adopt two classical image geometric transformations: flip and transpose.
To demonstrate their effectiveness, we individually remove the flip transformation (\emph{w/o F} for short), the transpose transformation (\emph{w/o T} for short), and both flip and transpose transformations (\emph{w/o F\&T} for short).
The comparison results in Table~\ref{tab: effect-of-scan} show that applying the flip or transpose transformation can achieve better results, improving the PSNR by at least $0.1$dB.
Our approach, which employs both flip and transpose transformations, outperforms all these baseline methods, with a notable improvement in PSNR by 0.26 dB, without significantly increasing the computational cost or runtime.
%

\vspace{-2mm}
{\flushleft \textbf{Effect of the efficient discriminative frequency domain-based FFN (EDFFN).}}
The EDFFN is mainly used to reduce computational complexity while maintaining performance.
To demonstrate its effectiveness, we compare two baseline methods that respectively only employ DFFN (\emph{w/ only DFFN} for short) and EDFFN (\emph{w/ only EDFFN} for short) in the encoder/decoder.
%
%
In contrast to DFFN that applies frequency-domain based operations in the middle of the FFN, where the number of channels of the feature is $6\times$ that of the input feature, our approach performs frequency-domain screening on the features at the final stage of the FFN network where the number of channels is the same as that of the input feature.
As Table~\ref{tab:EDFFN} shows, the runtime of the baseline with EDFFN is half that of the baseline with DFFN and the number of model parameters is reduced by 1.3M.
Concurrently, there is no loss in the performance of the baseline with EDFFN.
\begin{table}[!t]\normalsize
\centering
    \caption{Effectiveness of the EDFFN, evaluated on the GoPro dataset~\cite{GoPro}.}
    \vspace{-3mm}
    \label{tab:EDFFN}
    \footnotesize
    \setlength{\tabcolsep}{3pt} 
    \begin{tabular}{l|cccc} 
        \toprule
       Method         &Params          &GPU memory   &Runtime   &PSRN (dB)/SSIM  \\
\hline
        w/ only DFFN  & 11.5M           &  12.2G          & 75ms & 33.67/\textbf{0.9664}  \\
        w/ only EDFFN & \textbf{10.2M}  & \textbf{8.3G}   & \textbf{36ms} & \textbf{33.68}/0.9661 \\
        \bottomrule
    \end{tabular}
    \vspace{-2mm}

\end{table}
%

\begin{table}[!t]\footnotesize
\centering
    \caption{ Model complexity of the top-performance methods on the GoPro dataset, evaluated on images with the size of $256\times 256$ pixels. All the results are obtained on a machine with an NVIDIA RTX 3090 GPU.}
    \vspace{-3mm}

    \label{tab:runtime}
    \footnotesize
    \setlength{\tabcolsep}{1pt} 
    \begin{tabular}{l|ccccccccccccccc}
        \toprule
       Method  & IPT~\cite{IPT} &Restormer-local~\cite{TLC}  &GRL~\cite{GRL} &FFTformer~\cite{fftformer}            &EVSSM \\
\hline
        Params        &114M     & 26.1M      & 20.20M       & \textbf{16.6M}   &17.1M     \\
       FLOPs            &376G    &155G    &1289G          &131G   &\textbf{126G}\\
        Runtime        &298ms & 286ms   & 518ms    & 132ms          & \textbf{89ms}     \\

        \bottomrule
    \end{tabular}
    \vspace{-5mm}
\end{table}

\vspace{-2mm}
{\flushleft \textbf{Model complexity.}}
We further examine the model complexity of the proposed approach and other top-performance methods in terms of model parameters, floating point operations (FLOPs), and average runtime.
Table~\ref{tab:runtime} shows that the proposed method has fewer FLOPs and is faster
than the evaluated methods.

\vspace{-2mm}
{\flushleft \textbf{Limitations.}}
We develop an effective and efficient method that explores the properties of the state space model for high-quality image restoration.  
However, we have only considered simple transformations such as flip and transpose so far. 
In future work, we will consider more powerful transformation methods, such as the polar coordinate transformation, to better characterize the spatial information of visual data with SSMs.

\section{Conclusion}
In this paper, we propose an efficient visual state space model for image deblurring. 
Specifically, we develop an efficient visual scan block, where we employ various geometric transformations before each scan to adapt SSMs to visual data.
We show that compared to existing methods that scan along multiple directions simultaneously, our method is more effective at exploring non-local information without significantly increasing the computational cost.
Extensive evaluations and comparisons with state-of-the-art methods demonstrate that our approach is much more efficient while achieving favorable performance.

\noindent{\bf Acknowledgments.}
This work has been supported in part by the National Natural Science Foundation of China (Nos. 62272233, U22B2049, and 62332010).
M.-H. Yang is supported in part by the Institute of Information \& Communications Technology Planning \& Evaluation (IITP) grant funded by the Korean Government (MSIT) (No. RS-2024-00457882, National AI Research Lab Project).

{
    \small
    \bibliographystyle{ieeenat_fullname}
    \bibliography{main}
}


\end{document}